\documentclass[conference]{IEEEtran}
\IEEEoverridecommandlockouts
\usepackage{cite}
\usepackage{amsmath,amssymb,amsfonts}
\usepackage{algorithmic}
\usepackage{graphicx}
\usepackage{textcomp}
\usepackage{xcolor}
\usepackage{hyperref}
\usepackage{booktabs}
\usepackage{siunitx}
\usepackage{pifont}
\usepackage{booktabs}  
\usepackage{makecell}  
\usepackage{subcaption}
\usepackage{fmtcount}
\usepackage{geometry}

\def\BibTeX{{\rm B\kern-.05em{\sc i\kern-.025em b}\kern-.08em
    T\kern-.1667em\lower.7ex\hbox{E}\kern-.125emX}}

\begin{document}
% \newgeometry{
%  top=2.6cm,      % 上边距（含页眉）
%  bottom=1.9cm,   % 下边距（含页脚）
%  left=1.9cm,     % 左边距（装订侧）
%  right=1.9cm,    % 右边距
%  headheight=1cm, % 页眉高度
%  footskip=1cm    % 页脚基线距离
% } % 首页特殊设置
\graphicspath{{Images/}}

\title{LogisticsVLN: Vision-Language Navigation \\For Low-Altitude Terminal Delivery  \\Based on Agentic UAVs
\thanks{This work is partly supported by the Science and Technology Development Fund, Macao SAR (File no. 0145/2023/RIA3 and 0093/2023/RIA2), the National Natural Science Foundation of China (62303460), the Young Elite Scientists Sponsorship Program of China Association of Science and Technologyunder Grant YESS20220372. }
\thanks{Xinyuan Zhang is with the School of Artificial Intelligence, University of Chinese Academy of Sciences, Beijing 100049, China (e-mail: zhangxinyuan23@mails.ucas.ac.cn).}
\thanks{Yonglin Tian and Yue Liu are with the State Key Laboratory of Multimodal Artificial Intelligence Systems, Institute of Automation, Chinese Academy of Sciences, Beijing 100190, China (e-mail: yonglin.tian@ia.ac.cn, liuyue@2023@ia.ac.cn).} 
\thanks{Fei Lin is with the Department of Engineering Science, Faculty of Innovation Engineering, Macau University of Science and Technology, Macau 999078, China (e-mail: feilin@ieee.org).}
\thanks{Jing Ma is with China Ship Research and Development Academy, Beijing 100101, China (e-mail: majing21st@yeah.net)}
\thanks{Kornélia Sára Szatmáry is with Obuda University, Hungary (e-amil: szatmary.sara@stud.uni-obuda.hu).}
\thanks{Fei-Yue Wang is with the State Key Laboratory for Management and Control of Complex Systems, Chinese Academy of Sciences, Beijing 100190, and also with the Department of Engineering Science, Faculty of Innovation Engineering, Macau University of Science and Technology, Macau 999078, China (e-mail: feiyue.wang@ia.ac.cn).}
}

% \author{\IEEEauthorblockN{Xinyuan Zhang\IEEEauthorrefmark{1},
% Yonglin Tian\IEEEauthorrefmark{2},
% Fei Lin\IEEEauthorrefmark{3}, 
% Yue Liu\IEEEauthorrefmark{4}, and
% Fei-yue Wang\IEEEauthorrefmark{5}}
% \IEEEauthorblockA{\IEEEauthorrefmark{1}School of Artificial Intelligence,
% University of Chinese Academy of Sciences, \\
% Beijing, China, zhangxinyuan23@mails.ucas.ac.cn}
% \IEEEauthorblockA{\IEEEauthorrefmark{2}The State Key Laboratory of Multimodal Artificial Intelligence Systems, Institute of Automation, Chinese Academy of Sciences, \\
% Beijing, China, yonglin.tian@ia.ac.cn}
% \IEEEauthorblockA{\IEEEauthorrefmark{3}Department of Engineering Science, Faculty of Innovation Engineering, Macau University of Science and Technology, \\
% Macau, China, feilin@ieee.org}
% \IEEEauthorblockA{\IEEEauthorrefmark{4}Institute of Automation, Chinese Academy of Sciences, Beijing, China, liuyue@2023@ia.ac.cn}
% \IEEEauthorblockA{\IEEEauthorrefmark{5}State Key Laboratory for Management, Chinese Academy of Sciences, Beijing, China, feiyue@ieee.org}
% }
\author{Xinyuan Zhang, Yonglin Tian, Fei Lin, Yue Liu, Jing Ma, Kornélia Sára Szatmáry, Fei-Yue Wang}

\maketitle      % 标题页内容

\begin{abstract}
The growing demand for intelligent logistics, particularly fine-grained terminal delivery, underscores the need for autonomous UAV (Unmanned Aerial Vehicle)-based delivery systems. However, most existing last-mile delivery studies rely on ground robots, while current UAV-based Vision-Language Navigation (VLN) tasks primarily focus on coarse-grained, long-range goals, making them unsuitable for precise terminal delivery.  To bridge this gap, we propose LogisticsVLN, a scalable aerial delivery system built on multimodal large language models (MLLMs) for autonomous terminal delivery. LogisticsVLN integrates lightweight Large Language Models (LLMs) and Visual-Language Models (VLMs) in a modular pipeline for request understanding, floor localization, object detection, and action-decision making. To support research and evaluation in this new setting, we construct the Vision-Language Delivery (VLD) dataset within the CARLA simulator. Experimental results on the VLD dataset showcase the feasibility of the LogisticsVLN system. In addition, we conduct subtask-level evaluations of each module of our system, offering valuable insights for improving the robustness and real-world deployment of foundation model-based vision-language delivery systems. 
\end{abstract}


\section{Introduction}
Driven by the rapid growth of e-commerce and urbanization, logistics has become an increasingly critical component of modern society\cite{logistics1}. In particular, there is a growing demand for stable, efficient, and user-centric terminal delivery, which refers to the final step of transporting goods directly to the end user's residence\cite{mircrologistics}. One promising solution lies in applying agentic UAVs (Unmanned Aerial Vehicles)\cite{uavmeetllm}---UAV systems empowered by foundation models---to perform Vision-Language Navigation (VLN) in terminal delivery scenarios.  

Traditional VLN studies often rely on network-based approaches\cite{nomad}\cite{vint}, which typically require extensive training data to generalize. Most existing UAV-based VLN benchmarks\cite{openfly} \cite{openuav}\cite{uavvla} focus on long-range navigation with coarse-grained goals, making them unsuitable for terminal delivery scenarios. The OPEN system \cite{openbench} has paid attention to the last-mile delivery tasks. Following recent work on indoor VLN tasks\cite{navgpt}\cite{mapgpt}, OPEN adopts foundation models in understanding, perceiving, and planning, showing excellent zero-shot performance without training. However, OPEN is implemented on ground robots, which limits the delivery to the building level rather than a more refined window level. 

\begin{figure*}[t] % [t] 表示优先放在页面顶部
    \centering
    \includegraphics[width=\textwidth]{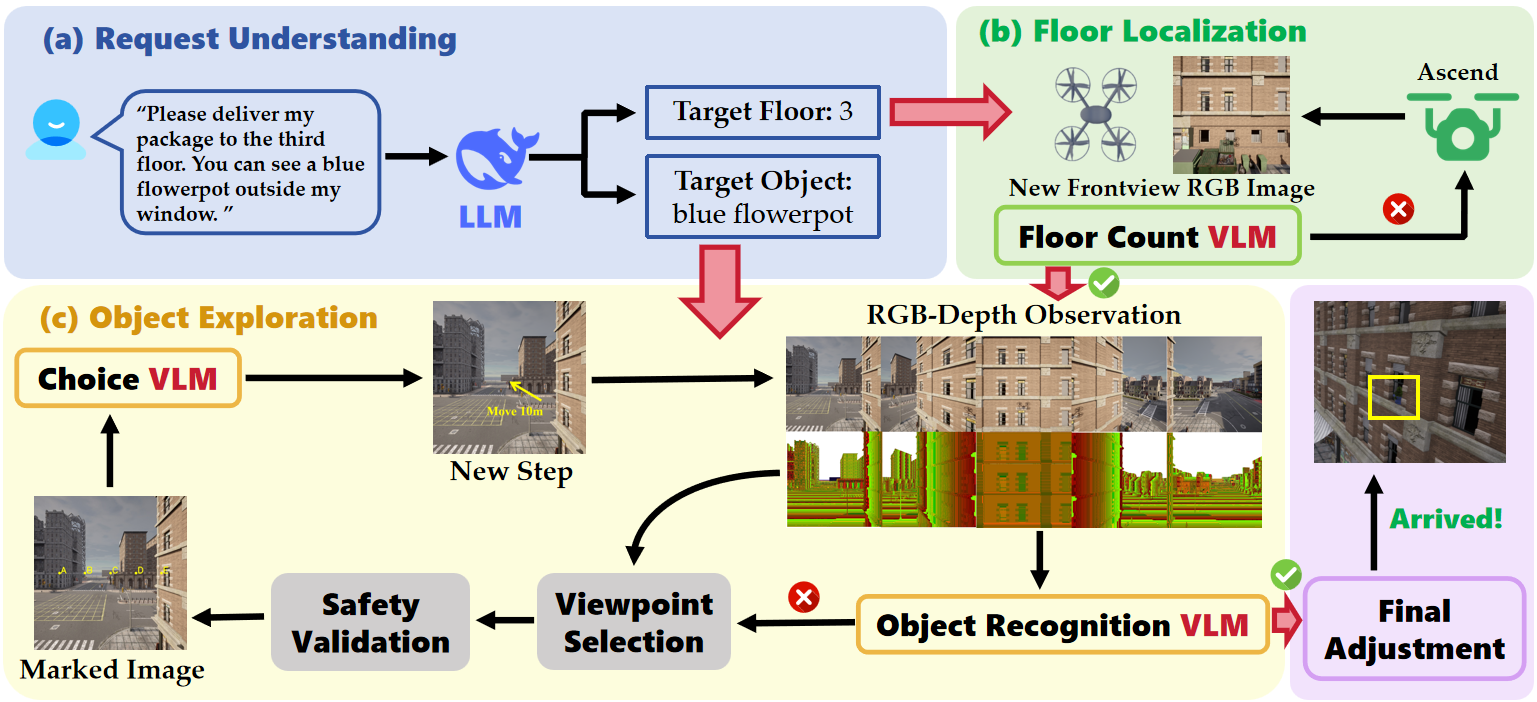} % 图片宽度与整个页面同宽
    \caption{Overview of the LogisticsVLN system based on an agentic UAV}
    \label{fig:overview}
\end{figure*}

To overcome these challenges, we propose LogisticsVLN (Fig.~\ref{fig:overview}), a UAV-based VLN system built on lightweight multimodal large language models (MLLMs). Our system provides a scalable solution for window-level terminal delivery tasks. 

The proposed system begins by leveraging a Large Language Model (LLM) to interpret customer requests and extract key attributes of the target window.  A floor localization method based on a Floor Count Visual-Language Model (VLM) is then employed to estimate floors and guide the drone to ascend to an appropriate height. Upon arriving at the target floor, the drone explores around the building in search of the customer's window with the help of a Viewpoint Selection algorithm, an Object Detection VLM, and a Choice VLM. A depth assistant module is also integrated to ensure operational safety. 

The main contributions of this work can be summarized as follows:
\begin{itemize}
\item We propose LogisticsVLN, the first aerial VLN system designed for window-level terminal delivery scenarios using only simple sensors and lightweight large models. The system requires no prior environmental knowledge or fine-tuning, making it highly deployable in unseen settings.
\item We construct a Vison-Language Delivery (VLD) dataset, 
% \newpage
% \restoregeometry
which focuses on continuous aerial scenarios for terminal delivery, filling a gap in existing VLN benchmarks. Built within the CARLA simulator, the dataset offers diverse scenarios and tasks for the evaluation of last-mile drone delivery systems. 
\item We apply MLLMs in the aerial delivery context and evaluate their roles and limitations in each subtask. We also analyze representative failure cases to guide future improvements and provide insights for deploying foundation models in real-world VLD applications.
\end{itemize}

% 跨双栏的大图

\section{Related Works}

\subsection{Visual-Language Navigation}

In recent years, VLN has emerged as a key task in embodied AI, requiring agents to interpret natural language instructions and perform corresponding navigation in complex environments. Most VLN tasks are built on simulators such as Matterport3D\cite{matterport} and AirSim\cite{airsim}, which provide realistic and cost-effective platforms for training and testing. Early work by Anderson et al.\cite{r2r} introduced the Room-to-Room (R2R) benchmark, where an agent needs to execute discrete actions to reach a designated goal following step-by-step instructions in indoor scenarios.  Krantz et al. \cite{vln-ce} extended this to continuous environments with VLN-CE, enabling more dynamic navigation. Tasks like REVERIE\cite{reverie} and SOON\cite{soon} adopt target-oriented instructions that more closely reflect how humans communicate. In these settings, the agent must understand both the high-level guidance and the environment semantically to locate and identify the specified targets. 

With the expansion of application scenarios, some research efforts have turned to outdoor navigation. On the ground, LM-Nav\cite{lmnav} combines three independent pretrained models to enable long-horizon instruction following in complex, real-world settings. In the aerial domain, Aerial VLN\cite{aerialvln} explores discrete drone navigation guided by natural language instructions, OpenUAV\cite{openuav} investigates continuous, goal-driven UAV navigation, and CityNav\cite{citynav} focuses on aerial navigation with geographic information. 

However, most existing outdoor UAV navigation research \cite{openfly}\cite{uavvla} centers on long-range missions with large-scale targets and broad navigation zones. In contrast, aerial navigation with small targets over short distances—especially in terminal delivery contexts—remains underexplored.

\subsection{MLLM-based UAV systems}
The integration of MLLMs can significantly enhance UAVs’ capabilities in perception, decision-making, and human-agent interaction. Limberg et al.\cite{lim-perception} attempted to apply YOLO-World and GPT-4V for zero-shot human detection and action recognition in UAV imagery. Kim et al.\cite{kim-perception} employed LLaVA-1.5 and CLIP encoder to interpret environmental factors in drone-view images, enhancing object recognition under challenging weather conditions of UAVs. Recent works like \cite{airvista2}  focus on improving UAVs' semantic understanding and reasoning capabilities in complex dynamic contents. 

Furthermore, AeroAgent\cite{aeroagent} enhances agents’ contextual understanding and multitask capabilities by integrating GPT-4V with a retrievable multimodal memory system. In the PMA framework \cite{cityeqa},  a hierarchical Planner-Manager-Actor architecture is designed, which combines pretrained LLMs for planning and MLLMs for visual perception in outdoor EQA tasks. 
% Patrol Agent \cite{patrolagent} employs VLMs for object detection and an on-cloud LLM for action-deciding, achieving autonomous urban operations. 
AirVista\cite{airvista}, proposed by Lin et al., incorporates ACP method with an MLLM agent to achieve autonomous flight and human-machine interaction. 

NavAgent\cite{navagent} applies MLLMs to drone VLN tasks. It uses GPT-4 for landmark text extraction, a fine-tuned GLIP model for visual recognition, and LLaMa2-13B for high-level decision-making, demonstrating strong performance in neighborhood navigation.   

These studies not only confirm the potential of MLLMs in enhancing operational intelligence of UAVs but also underscore the feasibility of building more generalizable, intelligent drone agents leveraging foundation models.
\subsection{Autonomous delivery}
Autonomous logistics delivery, especially terminal delivery tasks, holds great promise for reducing labor costs and improving service efficiency, making it a key area of research and industry. 

To address this domain, Wang et al. proposed OPEN system\cite{openbench}, which integrates foundation models with classic algorithms and uses off-the shelf OpenStreetMap to enable last-mile delivery on ground vehicles. 

Another work by Tian et al.\cite{logisticvista} introduced the LogisticsVISTA framework and its integrated simulator Logistics-Sim. This system focuses on efficient 3D terminal delivery by coordinating UAVs, UGVs (Unmanned Ground Vehicles), and USVs (Unmanned Surface Vehicles), while also incorporating LLMs and VLMs for user interaction and tailored services.  

Our proposed LogisticsVLN system is inspired by LogisticsVISTA and can serve as the implementation of its UAV-based terminal delivery part.

\section{LogisticsVLN}
% 本节分几个小节介绍方法
\subsection{Task Definition}
Our aerial autonomous terminal delivery task focuses on the final stage of logistics—the precise, window-level delivery task in residential environments. Starting from a position near the target building, the drone must reach the specific window of the user based solely on a natural language request, without relying on any pre-constructed maps. Considering that detailed internal maps are often unavailable in real-world residential areas, this setting closely reflects the practical challenges of real-world UAV-based terminal delivery.In summary, the goal is to formulate a policy \( \pi \) that maps the drone's observations \( \mathit{O} \) at each time step \( t \) to an action \( a_t \):

\begin{equation}
A_t = \pi(O_t, \mathcal{L})
\end{equation}

The policy is expected to produce a sequence of actions \( \{A_0, A_1, \dots, A_T\} \) such that the final state of the drone \( x_T \) satisfies:

\begin{equation}
x_T \in \mathcal{N}(w^*)
\end{equation}

where \( w^* \) is the inferred target window position based on \( \mathcal{R} \), and \( \mathcal{N}(w^*) \) denotes a small neighborhood around \( w^* \) representing successful delivery.

\subsection{System Overview}
An overview of the agentic UAV system we proposed for the defined delivery task is shown in Fig.~\ref{fig:overview}. The drone is equipped with five pairs of RGB-Depth cameras, oriented at \( R = \{0^\circ, 45^\circ, 90^\circ, -45^\circ, 90^\circ\} \), to capture RGB images \( \{I^{RGB}_{t,r_i} \}_{i=1}^{5} \) and depth images \( \{I^{Depth}_{t,r_i} \}_{i=1}^{5} \) at each time step \( t \). This arrangement enables a semi-panoramic perception of the surrounding environment.

The delivery process begins with a natural language request, which is interpreted by a Request Understanding module built upon an LLM. Guided by the information extracted from the request, two VLM-based modules---Floor Localization and Object Recognition---interact with the environment to identify the specific window \( w^* \) mentioned by the user. Specifically, three VLMs are employed across these two modules to handle floor estimation, target identification, and action decision, respectively. A depth assistant is also added to enhance perception by parsing spatial cues from depth observations (corresponding to the grey blocks in Fig.~\ref{fig:overview}). Once the target window is detected, the UAV adjusts its position under the guidance of the VLMs and approaches  \( w^* \) to complete the delivery.

\subsection{Request Understanding}
The request provided by the user in the terminal delivery task contains a description of the target window, including the item to be delivered, the floor location, distinctive surrounding objects (e.g., a green flower pot), and irrelevant or distracting details.

As illustrated in Fig.~\ref{fig:overview}(a), we leverage DeepSeek-R1-Distill-Qwen-14B\cite{deepseek} and design a few-shot prompt with three-step Chain-of-Thought (CoT) reasoning\cite{cot} to parse the input request. This process filters out irrelevant content and extracts two key elements: the target floor number\(F_{tar}\) and the distinctive surrounding object \(D_{tar}\). These outputs are then passed to subsequent modules for grounding and decision-making. 

\subsection{Floor Localization}
The Floor Localization module guides the drone to the correct floor where the \( w^* \) is located, which is made possible by a Floor Count VLM as shown in Fig.~\ref{fig:overview}(b). 

Starting from the base of the building, the drone generates a series of non-overlapping vertical waypoints \(\mathcal{H} = \{h_1, h_2, ...\}\) based on the Vertical Field of View (VFoV) of its front-facing camera. At each waypoint \(h_i\), the drone captures an RGB image \(I^{RGB}_{h_i, front}\), which is then analyzed by the Floor Count VLM to infer the number of floors visible \(F_{new,i}\) and update the current estimated position \(F_{cur}\) of the drone.

By comparing \(F_{cur}\) and \(F_{tar}\), the Floor Count VLM decides whether the drone should ascend to the next waypoint \(h_{i+1}\) or perform fine-grained adjustments to reach the desired floor height, which can be illustrated as follows: 
\[h_{\text{next}} =
\begin{cases} 
h_i - \dfrac{h_i - h_{i-1}}{F_{\text{new},i}} \cdot (F_{\text{cur}} - F_{\text{tar}}), & \text{if } F_{\text{cur}} \geq F_{\text{tar}}, \\
h_{i+1}, & \text{else}.
\end{cases}\tag{3}\]

Upon reaching the final height \(h_{\text{final}}\), the drone captures an RGB image and queries the VLM for the building's bounding box \(bbox_{\text{buil}} = [x_{\min}, x_{\max}, y_{\min}, y_{\max}]\). After this stage, the drone maintains a fixed altitude until \(D_{tar}\) is detected. 

\subsection{Object Exploration}
In the absence of pre-constructed maps, enabling the UAV to safely and efficiently explore objects around the building presents a significant challenge. Inspired by \cite{wmnav}, we propose a design composed of an object recognition VLM, a choice VLM, and a depth assistant, as depicted in Fig.~\ref{fig:overview}(c). This Object Exploration module processes all RGB-D images \( \{I^{RGB}_{t,r_i} \}_{i=1}^{5} \) and \( \{I^{Depth}_{t,r_i} \}_{i=1}^{5} \) captured at each time step \(t\). 

\textbf{1) Object Recognition: } \( \{I^{RGB}_{t,r_i} \}_{i=1}^{5} \) and \(D_{tar}\) are processed by the Object Recognition VLM to identify whether the target window is visible. If the VLM detects the target window in any RGB image, it returns the corresponding bounding box. The depth assistant then calculates a safe approach trajectory based on the drone's physical constraints, guiding the UAV to deliver the package precisely and securely.

\textbf{2) Viewpoint Selection: }If the Object Recognition VLM determines that the target window is not visible in \( \{I^{RGB}_{t,r_i} \}_{i=1}^{5} \), the Choice VLM and the depth assistant jointly determine the drone's next action. Since the terminal delivery scenario provides limited semantic information, constructing a curiosity value map purely based on VLM semantic understanding is ineffective. Therefore, we design a depth-based algorithm to calculate the score of each view and select the most promising one.

At each time step \( t \), for each visible camera view \( r_i \), we first cut off the upper and lower edges of the corresponding depth image \( I^{Depth}_{t,r_i} \) according to the \(bbox_{buil}\) given by the VLM in the final stage of floor localization.

Then, the cropped depth image is vertically partitioned into \( x \) equal-width slices. The average depth values for each slice are computed as:
\[
\mathcal{D}_{r_i} = \{ d_{r_i,1}, d_{r_i,2}, \ldots, d_{r_i,x} \},
\]
where \( d_{r_i,j} \) denotes the average depth value within the \( j \)-th vertical slice.

For each view \( r_i \), we search for a slice index \( j^*_{r_i} \) indicating a significant depth discontinuity. Specifically, for each candidate slice \( j \), we divide \( \mathcal{D}_{r_i} \) into a left partition \( \mathcal{D}_{r_i}^{\text{left}}(j) \) and a right partition \( \mathcal{D}_{r_i}^{\text{right}}(j) \). Assuming the drone is surrounding clockwise around the building, a valid split must satisfy:
\begin{itemize}
    \item The mean depth difference between left and right partitions is greater than a threshold \(\delta\):
    \[\text{mean}(\mathcal{D}_{r_i}^{left}(j)) - \text{mean}(\mathcal{D}_{r_i}^{right}(j)) \geq \delta
    \tag{4}\]
    \item The number of depth values in the right partition that exceed a maximum distance \(d_{\max}\) is limited.
\end{itemize}

Among all valid splits, we select the slice \( j^* \) that minimizes the total variance within the left and right partitions:
\[
j^* = \arg\min_j \left( \Sigma^2(\mathcal{D}_{r_i}^{left}(j)) + \Sigma^2(\mathcal{D}_{r_i}^{right}(j)) \right).
\tag{5}\]

The final exploration view \( r^* \) is selected as the one with the highest \( j^* \):
\[
r^* = \arg\max_{r_i} j^*_{r_i}.
\tag{6}\]

We select the right-facing camera as the default view and increase \(d_{max}\) when no valid \(j^*\) is found for any view.

\textbf{3) Action Selection: }At each time step \(t\), the action of the drone consists of its moving direction and moving distance, denoted as \(A_t = \{a_{t}^{dir}, a_{t}^{dis}\}\). We employ a Choice VLM to determine \(A_t\) with the help of a depth assistant.

We first mark five evenly distributed points \( \{p_k \}_{k=1}^{5} \) along the horizontal line at the center of the selected RGB image \(I^{RGB}_{r^*}\). For each marked point, the depth assistant calculates the safe moving distance \[l_k = \min\left(I^{\mathrm{depth}}(p_k), \, L_{\max}\right),\tag{7}\] with $L_{\max}$ being the maximum allowed distance per step. 

The marked image, along with the set of computed moving distances and the description of the delivery task, is formatted into prompts and provided to the Choice VLM. The Choice VLM selects one optimal point \(k\) as the intended moving direction \(a_{t}^{dir}\). The corresponding \(l_k\) is then directly taken as \(a_{t}^{dis}\). This mechanism enables the drone to navigate efficiently towards the next exploring spot while reducing the risk of collision.

\section{Experiments}

\subsection{VLD Dataset}

We construct a Visual-Language Delivery (VLD) Dataset using CARLA 0.9.12\cite{carla}, a highly realistic 3D simulation platform based on Unreal Engine 4. The built-in maps of CARLA cover urban, residential, and rural environments, providing a strong foundation for the diversity of our dataset.

\begin{figure}[htp]
    \centering
    \includegraphics[width=8cm]{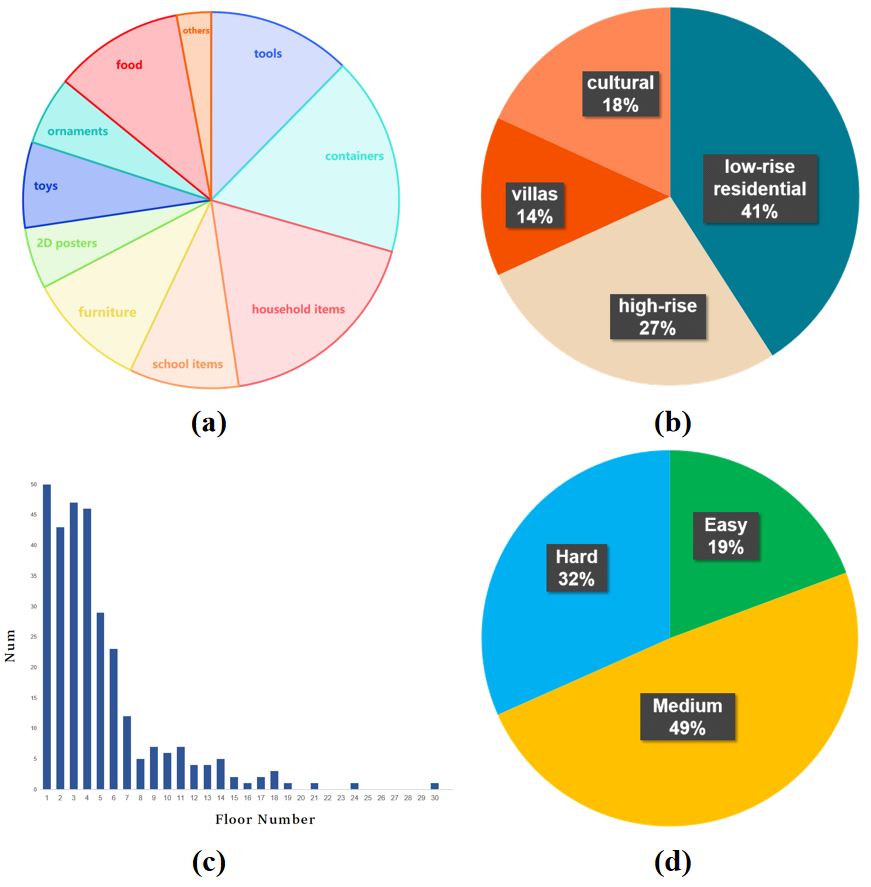}
    \caption{Statistics of VLD Dataset: (a) target categories, (b) building types, (c) floor number distribution, and (d) difficulty distribution. }
    \label{dataset}
\end{figure}

Based on both the predefined buildings and objects in CARLA maps and additional high-fidelity models manually added, we create 300 VLD tasks distributed across 22 different buildings. As shown in Fig.~\ref{dataset}(a), the target objects \(D_{tar}\) cover a wide range of categories, including tools, containers, household items, food, furniture, 2D posters, toys, and ornaments. The types of buildings include low-rise residential buildings, high-rise structures, small villas, and culturally themed architectures, which are depicted in Fig.~\ref{dataset}(b).

Besides the rich variety in scenes, we also ensure broad coverage of task difficulty levels and target floor numbers \(F_{tar}\), as shown in Fig~\ref{dataset}(c) and (d). The task difficulty is defined by the minimum number of drone turns required to detect the \(D_{tar}\). We classify tasks with fewer than 2 turns as 'Easy', those requiring 2 to 3 turns as 'Moderate', and those with more than 3 turns as 'Hard'. 

We also use an LLM to enhance the linguistic diversity of user requests. Given the \(D_{tar}\) and \(F_{tar}\), GPT-4o\cite{gpt-4o} is prompted to simulate customer interactions and generate various request sentences with different tones and expression styles. These generated requests are then reviewed and refined by human experts to ensure data quality.

\subsection{Evaluation Metrics and Implementation Details}

\textbf{Evaluation Metrics:} In the VLD tasks, we adopt two widely used metrics for evaluating VLN performance: Success Rate (SR) and Success Weighted by Path Length (SPL)\cite{sp&spl}. SR measures the percentage of successfully completed tasks, while SPL evaluates both task completion and navigation efficiency.

Considering that the drone can only navigate around buildings without passing through them, we define the shortest path distance in SPL as the minimal distance required for the drone to fly from the starting point to the goal point along straight-line segments connecting the building's vertices.

We also add \textit{Average Steps} as an evaluation metric, which refers to the number of times the drone stops to capture an image and interact with the VLMs. This metric can reflect the time and memory footprint required to complete a VLD task. Only successful tasks are included in the calculation of Average Steps.

\textbf{Implementation Details:} In our experiment, we use a quadrotor UAV model as the drone in the simulator. All the cameras on the drone have a resolution of 800 × 800 pixels and a FoV of \(90^\circ\). All cameras are mounted beneath the drone to prevent the drone body from appearing in the captured images and interfering with perception. We limited the total number of time steps for exploration to 30 and set the slice number \(x\) in the viewpoint selection algorithm to 20. For request understanding, we employ DeepSeek-R1-Distill-Qwen-14B as the LLM. 
\subsection{Results}

We evaluate our VLD system on all 300 tasks using three lightweight VLMs: Qwen2-VL-7B-Instruct\cite{qwen2-vl}, LLaMA-3.1-11B-Vision-Instruct\cite{llama3}, and Yi-VL-6B\cite{yi}. The quantitative results are presented in Table~\ref{tab:vlm}.
%Notably, the same VLM is responsible for all vision-language-related subtasks.  

With Qwen2-VL, our system is able to successfully complete over half of the tasks in the VLD dataset. However, LLaMA-3.1 shows moderately worse performance, and Yi-VL falls behind significantly. 

We make further analysis and identify two major sources of this difference: (1) the accuracy of object recognition, and (2) the reliability of floor localization.

We observe that Yi-VL frequently refuses to give a precise answer in the Floor Count module in over half of the tasks, although we have explicitly claimed that the Floor Count VLM must provide a numerical floor estimation in the text prompt. For LLaMA-3.1, a more detailed comparison of its performance in Floor Localization with Qwen2-VL is presented in row a and b of Table~\ref{tab:floor localization}. Given that a drone image often spans multiple floors, our system allows some tolerance in localization. Therefore, we define a floor localization failure as a case where the absolute difference between the final stopped height \( h_{final} \) and the ground-truth target height \( h_{tar} \) exceeds 7m in the CARLA simulator. Based on this criterion, Llama-3.1 underperforms compared to Qwen2-VL in floor localization accuracy. 

The comparison of Object Recognition failure rate, which is defined as the ratio between the number of tasks where the VLM incorrectly identifies the target object and the total number of tasks where the VLM claims to have located the target, is shown in Table~\ref{tab:object recognition}. Compared to Qwen2-VL, Llama-3.1 and Yi-VL show a higher rate of misidentification, which is often due to over-reliance on attributes, especially color, mentioned in the extracted \(D_{tar}\). This can lead the VLM to incorrectly match non-target objects that share similar visual traits with the described object, resulting in the failure of delivery tasks. 

\begin{table}[t]
\centering

\caption{Evaluation Result with Different VLMs}
\label{tab:vlm}
\begin{tabular}{
  c  % 第一列居中（Choice VLM）
  S[table-format=2.1]  % SR列（2位整数+1位小数）
  S[table-format=2.1]  % SPL列
  S[table-format=2.3]  % Steps列
}
\toprule
\textbf{VLM} & \textbf{SR (\%) $\uparrow$} & \textbf{SPL (\%) $\uparrow$} & \textbf{Average Steps $\downarrow$} \\
\midrule
\makecell{Qwen2-VL-7B} & 54.7 & 50.8 & 12.15 \\
\makecell{Llama-3.1-11B} & 38.4 & 36.4 & 11.69 \\
\makecell{Yi-VL-6B} & 13.2 & 12.0 & 12.22 \\
\bottomrule
\end{tabular}
\end{table}

\begin{table}[t]
\centering
\caption{Evaluation on Floor Localization}
\label{tab:floor localization}  
\begin{tabular}{
  c  % 编号列
  c  % VLM 模型列
  c  % FL Method 列
  S[table-format=2.1]  % FL Fail Rate 列
}
\toprule
\textbf{ID} & \textbf{VLM} & \textbf{FL Method} & \textbf{FL Fail Rate (\%) $\downarrow$} \\
\midrule
a & Qwen2-VL-7B & Ours & 24.9 \\
b & Llama-3.1-11B & Ours & 27.1 \\
c & Qwen2-VL-7B & Direct Count & 61.6 \\
\bottomrule
\end{tabular}
\end{table}

\begin{table}[t]
\centering
\caption{Evaluation of Object Recognition}
\label{tab:object recognition}
\begin{tabular}{
  c  % 第一列居中（Choice VLM）
  S[table-format=2.1] 
}
\toprule
\textbf{VLM} & \textbf{OR Fail Rate (\%) $\downarrow$}\\
\midrule
\makecell{Qwen2-VL-7B} & 15.9  \\
\makecell{Llama-3.1-11B} & 40.4 \\
\makecell{Yi-VL-6B} & 53.3  \\
\bottomrule
\end{tabular}
\end{table}

\subsection{Ablation Studies}

\textbf{Effect of Our Floor Localization Method} As highlighted in the previous analysis, VLMs show deficiencies in floor counting reliability. Therefore, we conducted an ablation study to investigate the contribution of our Floor Localization Method.

We compare our proposed method with a \textit{Direct Count} method, in which the VLM is provided with a side-view image of the whole building and prompted to estimate its total number of floors. The assumption of uniform floor height and the definition of floor localization failure remain the same as previously stated.

Row a and c in Table~\ref{tab:floor localization} present the results. Our Floor Localization Method decreases the floor localization failure rate from 61.6\% to 27.9\%, demonstrating its effectiveness in enhancing floor localization performance.

\textbf{Effect of the Viewpoint Selection Algorithm} We further compare our proposed Viewpoint Selection algorithm with two alternative strategies. The \textit{Random} strategy selects one viewpoint randomly from \(\{r_i\}_{i=1}^5\), while the \textit{Default} strategy always selects the right-facing viewpoint under all circumstances.

Apart from the selection strategy, all other procedures remain consistent with our original method. We evaluate the performance of each strategy using Qwen2-VL on 50 randomly sampled tasks.

As presented in Table~\ref{tab:vs ablation}, both alternative viewpoint selection methods lead to noticeable drops in SR and SPL, showing the effectiveness of our proposed algorithm. This can be attributed to the fact that these strategies only succeed 'Easy' tasks where the target object is directly visible without requiring any viewpoint adjustment. This also explains their lower Average Steps value and identical SR and SPL rates.

\begin{table}[t]
\centering
\caption{Ablation Study on Viewpoint Selection Algorithm}
\label{tab:vs ablation}
\begin{tabular}{
  c  % 第一列居中（VS Strategy）
  S[table-format=2.1]  % SR列（2位整数+1位小数）
  S[table-format=2.1]  % SPL列
  S[table-format=2.2]  % Steps列（2位整数+2位小数）
}
\toprule
\textbf{VS Strategy}  & \textbf{SR (\%) $\uparrow$} & \textbf{SPL (\%) $\uparrow$} & \textbf{Average Steps $\downarrow$} \\
\midrule
\makecell {Ours} & 54.0 & 52.4 & 12.84\\
\makecell {Random} & 24.0& 24.0& 8.42  \\
\makecell {Default} & 20.0 & 20.0 & 8.80 \\
\bottomrule
\end{tabular}
\end{table}

\textbf{Effect of Choice VLM} Table~\ref{tab:choice ablation} illustrates the ablation study aiming at investigating the impact of the Choice VLM. The experiment is conducted on 200 randomly selected tasks, and removing Choice VLM implies always selecting the center point of the chosen viewpoint. Results show that the performance metrics across these two settings exhibit only minor differences.

One possible explanation lies in the fallback mechanism our system incorporated: if the depth assistant determines that the safe moving distance in the chosen direction at time step \(t\) is below a threshold, the drone performs a 30° left rotation to avoid reaching a deadlock. This mechanism ensures that if moving forward is not viable at time \(t\), it often becomes an efficient choice after such orientation adjustment at time \(t+1\).

\begin{figure}[htp]
    \centering
    \includegraphics[width=8cm]{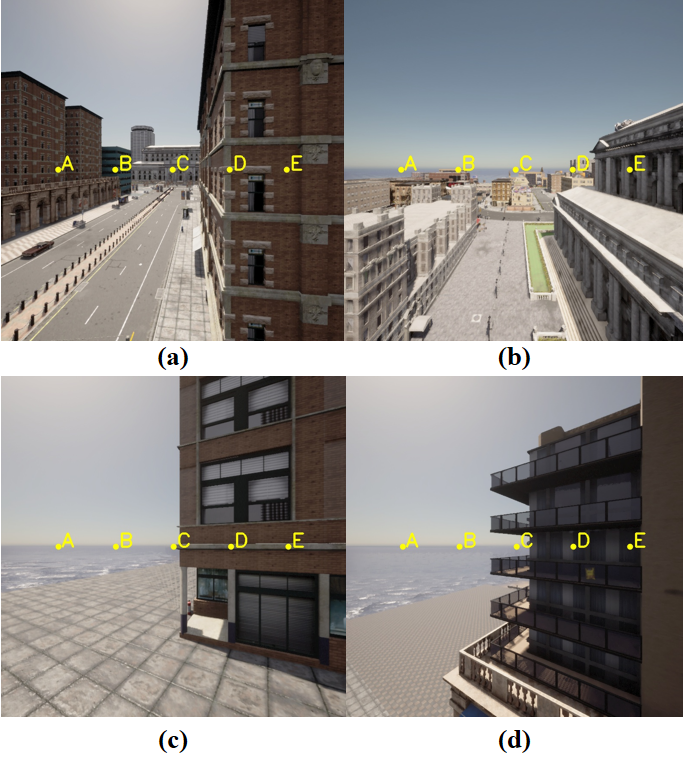}
    \caption{Examples and case studies about the effect of Choice VLM: (a) and (b) are examples of ideal viewpoint selection, while (c) and (d) are two case studies of suboptimal viewpoint.}
    \label{example}
\end{figure}

Another explanation may be the limited directional variation among the five marked points in \(I^{RGB}_{r_i}\) once a suitable viewpoint \(r_i\) is selected. Moreover, the Viewpoint Selection Algorithm presents a high accuracy in aligning the drone with the building's edge. In fact, Choice VLM ends up selecting the middle point and the point immediately to the left of the middle in over 70\% of cases. As illustrated in Fig.~\ref{example}(a) and (b), moving forward is the optimal action to continue searching around the building when the viewpoint is selected perfectly. This further explains the minimal performance gap observed in Table~\ref{tab:choice ablation}. 

When the viewpoint selection is suboptimal, the role of the Choice VLM becomes more indispensable. This can be illustrated by two real cases extracted from the drone’s execution process).

In Fig.~\ref{example}(c), the Choice VLM instructs the drone to move in direction b. This effective decision leads the drone to an unexplored side of the building. In contrast, choosing the default forward direction c would yield no new visual information, wasting valuable time and memory resources.

A more critical task built on a cultural architecture is shown in Fig.~\ref{example}(d).  Due to the intricate architectural design, the depth assistant fails to estimate a safe moving distance for direction c. Therefore, the drone would collide with the structure following the default forward direction. However, the Choice VLM correctly selects direction b, which enables the drone to avoid the accident and continue to explore safely.
\begin{table}[t]
\centering
\caption{Ablation Study on Choice VLM}
\label{tab:choice ablation}
\begin{tabular}{
  c  % 第一列居中（Choice VLM）
  S[table-format=2.1]  % SR列（2位整数+1位小数）
  S[table-format=2.1]  % SPL列
  S[table-format=2.2]  % Steps列（2位整数+2位小数）
}
\toprule
\textbf{Choice VLM}  & \textbf{SR (\%) $\uparrow$} & \textbf{SPL (\%) $\uparrow$} & \textbf{Steps $\downarrow$} \\
\midrule
\ding{51} & 52.9 & 50.4 & 11.68 \\
\ding{55} & 53.8 & 52.3 & 12.29 \\
\bottomrule
\end{tabular}
\end{table}

\section{Conclusion}
In this paper, we presented LogisticsVLN, a scalable UAV-based system for autonomous terminal delivery to building windows, leveraging the power of foundation models without task-specific training or pre-built maps. To facilitate evaluation in this new domain, we constructed the VLD dataset, which covers diverse building types, delivery targets, and instruction styles within the CARLA simulator. Experiments demonstrated the effectiveness of the proposed system, while subtask-level analyses provide valuable insights into the strengths and limitations of current VLMs in VLD scenarios. Future work will focus on optimizing system architecture to better exploit the capabilities of MLLMs and on extending LogisticsVLN to real-world aerial delivery applications.

\bibliographystyle{IEEEtran}
\bibliography{refs}

\begin{thebibliography}{10}
\providecommand{\url}[1]{#1}
\csname url@samestyle\endcsname
\providecommand{\newblock}{\relax}
\providecommand{\bibinfo}[2]{#2}
\providecommand{\BIBentrySTDinterwordspacing}{\spaceskip=0pt\relax}
\providecommand{\BIBentryALTinterwordstretchfactor}{4}
\providecommand{\BIBentryALTinterwordspacing}{\spaceskip=\fontdimen2\font plus
\BIBentryALTinterwordstretchfactor\fontdimen3\font minus \fontdimen4\font\relax}
\providecommand{\BIBforeignlanguage}[2]{{%
\expandafter\ifx\csname l@#1\endcsname\relax
\typeout{** WARNING: IEEEtran.bst: No hyphenation pattern has been}%
\typeout{** loaded for the language `#1'. Using the pattern for}%
\typeout{** the default language instead.}%
\else
\language=\csname l@#1\endcsname
\fi
#2}}
\providecommand{\BIBdecl}{\relax}
\BIBdecl

\bibitem{logistics1}
S.~Liu, X.~Zhuang, L.~Yan, Y.~Wang, S.~Wu, Y.~Lv, F.~Zhu, and F.-Y. Wang, ``A parallel logistic network simulation method and system to improve logistics efficiency,'' \emph{IEEE Journal of Radio Frequency Identification}, vol.~8, pp. 580--591, 2024.

\bibitem{mircrologistics}
J.~Havenga, ``Logistics and the future: The rise of macrologistics,'' \emph{Journal of Transport and Supply Chain Management}, vol.~12, 05 2018.

\bibitem{uavmeetllm}
Y.~{Tian}, F.~{Lin}, Y.~{Li}, T.~{Zhang}, Q.~{Zhang}, X.~{Fu}, J.~{Huang}, X.~{Dai}, Y.~{Wang}, C.~{Tian}, B.~{Li}, Y.~{Lv}, L.~{Kov{\'a}cs}, and F.-Y. {Wang}, ``{UAVs Meet LLMs: Overviews and Perspectives Toward Agentic Low-Altitude Mobility},'' \emph{arXiv e-prints}, p. arXiv:2501.02341, Jan. 2025.

\bibitem{nomad}
A.~{Sridhar}, D.~{Shah}, C.~{Glossop}, and S.~{Levine}, ``{NoMaD: Goal Masked Diffusion Policies for Navigation and Exploration},'' \emph{arXiv e-prints}, p. arXiv:2310.07896, Oct. 2023.

\bibitem{vint}
D.~{Shah}, A.~{Sridhar}, {Dashora} \emph{et~al.}, ``{ViNT: A Foundation Model for Visual Navigation},'' \emph{arXiv e-prints}, p. arXiv:2306.14846, Jun. 2023.

\bibitem{openfly}
Y.~{Gao}, C.~{Li}, {You} \emph{et~al.}, ``{OpenFly: A Versatile Toolchain and Large-scale Benchmark for Aerial Vision-Language Navigation},'' \emph{arXiv e-prints}, p. arXiv:2502.18041, Feb. 2025.

\bibitem{openuav}
X.~{Wang}, D.~{Yang}, {Wang} \emph{et~al.}, ``{Towards Realistic UAV Vision-Language Navigation: Platform, Benchmark, and Methodology},'' \emph{arXiv e-prints}, p. arXiv:2410.07087, Oct. 2024.

\bibitem{uavvla}
O.~{Sautenkov}, Y.~{Yaqoot}, {Lykov} \emph{et~al.}, ``{UAV-VLA: Vision-Language-Action System for Large Scale Aerial Mission Generation},'' \emph{arXiv e-prints}, p. arXiv:2501.05014, Jan. 2025.

\bibitem{openbench}
J.~{Wang}, D.~{Huo}, Z.~{Xu}, {Shi} \emph{et~al.}, ``{OpenBench: A New Benchmark and Baseline for Semantic Navigation in Smart Logistics},'' \emph{arXiv e-prints}, p. arXiv:2502.09238, Feb. 2025.

\bibitem{navgpt}
G.~{Zhou}, Y.~{Hong}, and Q.~{Wu}, ``{NavGPT: Explicit Reasoning in Vision-and-Language Navigation with Large Language Models},'' \emph{arXiv e-prints}, p. arXiv:2305.16986, May 2023.

\bibitem{mapgpt}
J.~{Chen}, B.~{Lin}, {Xu} \emph{et~al.}, ``{MapGPT: Map-Guided Prompting with Adaptive Path Planning for Vision-and-Language Navigation},'' \emph{arXiv e-prints}, p. arXiv:2401.07314, Jan. 2024.

\bibitem{matterport}
A.~{Chang}, A.~{Dai}, T.~{Funkhouser}, M.~{Halber}, M.~{Nie{\ss}ner}, M.~{Savva}, S.~{Song}, A.~{Zeng}, and Y.~{Zhang}, ``{Matterport3D: Learning from RGB-D Data in Indoor Environments},'' \emph{arXiv e-prints}, p. arXiv:1709.06158, Sep. 2017.

\bibitem{airsim}
S.~{Shah}, D.~{Dey}, C.~{Lovett}, and A.~{Kapoor}, ``{AirSim: High-Fidelity Visual and Physical Simulation for Autonomous Vehicles},'' \emph{arXiv e-prints}, p. arXiv:1705.05065, May 2017.

\bibitem{r2r}
P.~{Anderson}, Q.~{Wu}, D.~{Teney}, J.~{Bruce}, M.~{Johnson}, N.~{S{\"u}nderhauf}, I.~{Reid}, S.~{Gould}, and A.~{van den Hengel}, ``{Vision-and-Language Navigation: Interpreting visually-grounded navigation instructions in real environments},'' \emph{arXiv e-prints}, p. arXiv:1711.07280, Nov. 2017.

\bibitem{vln-ce}
J.~{Krantz}, E.~{Wijmans}, A.~{Majumdar}, D.~{Batra}, and S.~{Lee}, ``{Beyond the Nav-Graph: Vision-and-Language Navigation in Continuous Environments},'' \emph{arXiv e-prints}, p. arXiv:2004.02857, Apr. 2020.

\bibitem{reverie}
Y.~{Qi}, Q.~{Wu}, P.~{Anderson}, X.~{Wang}, W.~Y. {Wang}, C.~{Shen}, and A.~{van den Hengel}, ``{REVERIE: Remote Embodied Visual Referring Expression in Real Indoor Environments},'' \emph{arXiv e-prints}, p. arXiv:1904.10151, Apr. 2019.

\bibitem{soon}
F.~{Zhu}, X.~{Liang}, Y.~{Zhu}, X.~{Chang}, and X.~{Liang}, ``{SOON: Scenario Oriented Object Navigation with Graph-based Exploration},'' \emph{arXiv e-prints}, p. arXiv:2103.17138, Mar. 2021.

\bibitem{lmnav}
D.~{Shah}, B.~{Osinski}, B.~{Ichter}, and S.~{Levine}, ``{LM-Nav: Robotic Navigation with Large Pre-Trained Models of Language, Vision, and Action},'' \emph{arXiv e-prints}, p. arXiv:2207.04429, Jul. 2022.

\bibitem{aerialvln}
S.~{Liu}, H.~{Zhang}, Y.~{Qi}, P.~{Wang}, Y.~{Zhang}, and Q.~{Wu}, ``{AerialVLN: Vision-and-Language Navigation for UAVs},'' \emph{arXiv e-prints}, p. arXiv:2308.06735, Aug. 2023.

\bibitem{citynav}
J.~{Lee}, T.~{Miyanishi}, S.~{Kurita}, K.~{Sakamoto}, D.~{Azuma}, Y.~{Matsuo}, and N.~{Inoue}, ``{CityNav: Language-Goal Aerial Navigation Dataset with Geographic Information},'' \emph{arXiv e-prints}, p. arXiv:2406.14240, Jun. 2024.

\bibitem{lim-perception}
C.~{Limberg}, A.~{Gon{\c{c}}alves}, B.~{Rigault}, and H.~{Prendinger}, ``{Leveraging YOLO-World and GPT-4V LMMs for Zero-Shot Person Detection and Action Recognition in Drone Imagery},'' \emph{arXiv e-prints}, p. arXiv:2404.01571, Apr. 2024.

\bibitem{kim-perception}
H.~Kim, D.~Lee, S.~Park, and Y.~M. Ro, ``Weather-aware drone-view object detection via environmental context understanding,'' in \emph{2024 IEEE International Conference on Image Processing (ICIP)}, 2024, pp. 549--555.

\bibitem{airvista2}
F.~{Lin}, Y.~{Tian}, T.~{Zhang}, J.~{Huang}, S.~{Guan}, and F.-Y. {Wang}, ``{AirVista-II: An Agentic System for Embodied UAVs Toward Dynamic Scene Semantic Understanding},'' \emph{arXiv e-prints}, p. arXiv:2504.09583, Apr. 2025.

\bibitem{aeroagent}
H.~{Zhao}, F.~{Pan}, H.~{Ping}, and Y.~{Zhou}, ``{Agent as Cerebrum, Controller as Cerebellum: Implementing an Embodied LMM-based Agent on Drones},'' \emph{arXiv e-prints}, p. arXiv:2311.15033, Nov. 2023.

\bibitem{cityeqa}
Y.~{Zhao}, K.~{Xu}, Z.~{Zhu}, Y.~{Hu}, Z.~{Zheng}, Y.~{Chen}, Y.~{Ji}, C.~{Gao}, Y.~{Li}, and J.~{Huang}, ``{CityEQA: A Hierarchical LLM Agent on Embodied Question Answering Benchmark in City Space},'' \emph{arXiv e-prints}, p. arXiv:2502.12532, Feb. 2025.

\bibitem{airvista}
F.~Lin, Y.~Tian, Y.~Wang, T.~Zhang, X.~Zhang, and F.-Y. Wang, ``Airvista: Empowering uavs with 3d spatial reasoning abilities through a multimodal large language model agent,'' in \emph{2024 IEEE 27th International Conference on Intelligent Transportation Systems (ITSC)}, 2024, pp. 476--481.

\bibitem{navagent}
Y.~{Liu}, F.~{Yao}, Y.~{Yue}, G.~{Xu}, X.~{Sun}, and K.~{Fu}, ``{NavAgent: Multi-scale Urban Street View Fusion For UAV Embodied Vision-and-Language Navigation},'' \emph{arXiv e-prints}, p. arXiv:2411.08579, Nov. 2024.

\bibitem{logisticvista}
Y.~Tian, F.~Lin, X.~Zhang, J.~Ge, Y.~Wang, X.~Dai, Y.~Lv, and F.-Y. Wang, ``Logisticsvista: 3d terminal delivery services with uavs, ugvs and usvs based on foundation models and scenarios engineering,'' in \emph{2024 IEEE International Conference on Service Operations and Logistics, and Informatics (SOLI)}, 2024, pp. 2--7.

\bibitem{deepseek}
{DeepSeek-AI}, ``{DeepSeek-R1: Incentivizing Reasoning Capability in LLMs via Reinforcement Learning},'' \emph{arXiv e-prints}, p. arXiv:2501.12948, Jan. 2025.

\bibitem{cot}
J.~{Wei}, X.~{Wang}, D.~{Schuurmans}, M.~{Bosma}, B.~{Ichter}, F.~{Xia}, E.~{Chi}, Q.~{Le}, and D.~{Zhou}, ``{Chain-of-Thought Prompting Elicits Reasoning in Large Language Models},'' \emph{arXiv e-prints}, p. arXiv:2201.11903, Jan. 2022.

\bibitem{wmnav}
D.~{Nie}, X.~{Guo}, Y.~{Duan}, R.~{Zhang}, and L.~{Chen}, ``{WMNav: Integrating Vision-Language Models into World Models for Object Goal Navigation},'' \emph{arXiv e-prints}, p. arXiv:2503.02247, Mar. 2025.

\bibitem{carla}
A.~{Dosovitskiy}, G.~{Ros}, F.~{Codevilla}, A.~{Lopez}, and V.~{Koltun}, ``{CARLA: An Open Urban Driving Simulator},'' \emph{arXiv e-prints}, p. arXiv:1711.03938, Nov. 2017.

\bibitem{gpt-4o}
{OpenAI}, ``{GPT-4o System Card},'' \emph{arXiv e-prints}, p. arXiv:2410.21276, Oct. 2024.

\bibitem{sp&spl}
P.~{Anderson}, A.~{Chang}, D.~{Singh Chaplot}, A.~{Dosovitskiy}, S.~{Gupta}, V.~{Koltun}, J.~{Kosecka}, J.~{Malik}, R.~{Mottaghi}, M.~{Savva}, and A.~R. {Zamir}, ``{On Evaluation of Embodied Navigation Agents},'' \emph{arXiv e-prints}, p. arXiv:1807.06757, Jul. 2018.

\bibitem{qwen2-vl}
P.~{Wang}, S.~{Bai}, {Tan} \emph{et~al.}, ``{Qwen2-VL: Enhancing Vision-Language Model's Perception of the World at Any Resolution},'' \emph{arXiv e-prints}, p. arXiv:2409.12191, Sep. 2024.

\bibitem{llama3}
A.~{Grattafiori}, A.~{Dubey}, {Jauhri} \emph{et~al.}, ``{The Llama 3 Herd of Models},'' \emph{arXiv e-prints}, p. arXiv:2407.21783, Jul. 2024.

\bibitem{yi}
01.AI, A.~Young, B.~Chen \emph{et~al.}, ``{Yi: Open Foundation Models by 01.AI},'' \emph{arXiv e-prints}, p. arXiv:2403.04652, Mar. 2024.

\end{thebibliography}
\vspace{12pt}

\end{document}